# A Vietnamese Question Answering System

Dai Quoc Nguyen, Dat Quoc Nguyen, Son Bao Pham
Human Machine Interaction Laboratory
Faculty of Information Technology
College of Technology
Vietnam National University, Hanoi
{dainq, datnq, sonpb}@vnu.edu.vn

*Abstract*—Question answering systems aim to produce exact answers to users' questions instead of a list of related documents as used by current search engines. In this paper, we propose an ontology-based Vietnamese question answering system that allows users to express their questions in natural language. To the best of our knowledge, this is the first attempt to enable users to query an ontological knowledge base using Vietnamese natural language. Experiments of our system on an organizational ontology show promising results.

## I. INTRODUCTION

The goal of question answering systems is to give exact answers to the user's questions instead of just a list of related documents as used by most current search engines [1]. It is also desirable to allow users to specify the question using natural language expressions rather than the keyword-based approach. This is an avenue that has not been actively explored for Vietnamese.

In this paper, we introduce a domain independent Vietnamese question answering system. The target domain is modeled as an ontology in our system to leverage techniques and recent advances in semantic web. Thus semantic markups can be used to add meta-information to provide precise answers to complex questions expressed in natural language.

Our system contains a front-end that performs syntactic and semantic analysis on natural language questions on GATE framework [7]. The back-end is responsible for making sense of the user query with respect to a target ontology using various concept-matching techniques between a natural language phrase and elements in the ontology. The communication between the front-end and back-end is an intermediate representation of the question, which captures the semantic structure of the users' query.

The rest of paper is designed as follows: in section II, we provide some related works and describe our system in section III. We demonstrate the complete working system in section IV and evaluate the system in section V. The conclusion and future works will be presented in section VI.

## II. RELATED WORKS

K. Nguyen and H. Le [9] introduce a NLIDB (Natural Language Interface to DataBases) question answering system in Vietnamese employing semantic grammars. Their system includes 2 main modules: QTRAN and TGEN. QTRAN (Query Translator) maps a natural language question to an SQL query while TGEN (Text Generator) generates answers based on the query result tables. QTRAN uses limited context-free grammars to analyze user's question into syntax tree via CYK algorithm. The syntax tree is then converted into an SQL query by using a mapping dictionary to determine names of attributes in Vietnamese, names of attributes in the database and names of individuals stored these attributes. TGEN module combines pattern-based and keyword-based approaches to make sense of the meta-data and relations in the database tables in order to generate semantic answers.

PRECISE [2] is a NLIDB question answering system that takes as its input a natural language question to generate a corresponding SQL query. PRECISE showed a high precision (over 80% in a list of hundreds English questions). However PRECISE requires all tokens in input questions to be distinct and appear in its lexicon.

Aqualog [14] is an ontology-based question answering system for English and is the basis for the development of our system. Aqualog takes a natural language question and an ontology as its input, and returns an answer for users based on the semantic analysis of the question and the corresponding elements in the ontology. Aqualog's architecture can be described as a waterfall model where a natural language question is mapped to a set of representation based on the intermediate triple that is called a Query-Triple through the Linguistic Component. The Relation Similarity Service takes a Query-Triple and processes it to provide queries with respect to the input ontology called Onto-Triple.

Aqualog performs semantic and syntactic analysis of the input question through the use of processing resources provided by GATE [7] such as word segmentation, sentence segment, part-of-speech tagging. When a question is asked, the task of Linguistic Component is to transfer the natural language question to a Query-Triple with the following format (generic term, relation, second term). Through the use of JAPE grammars in GATE, AquaLog identifies terms and their relationship. The Relation Similarity Service uses Query-Triples to create Ontology-Triples where each term in the Query-Triples is matched with elements in the ontology.

QuestIO is another ontology-based question answering system using the GATE framework to analyze natural language questions [6].

A. Galea [1] introduced an open-domain question answering system using the GATE framework to process natural language questions.

### III. OUR SYSTEM ARCHITECTURE

Following Aqualog [14], our system is implemented in Java as a web application using a client-server architecture. The general architecture of our question answering system is shown in Figure 1. It includes two components: the natural language question analysis engine and the answer retrieval module. The question analysis component maps the question into an intermediate representation tuple, which is then fed into the answer retrieval component to generate a semantic answer.

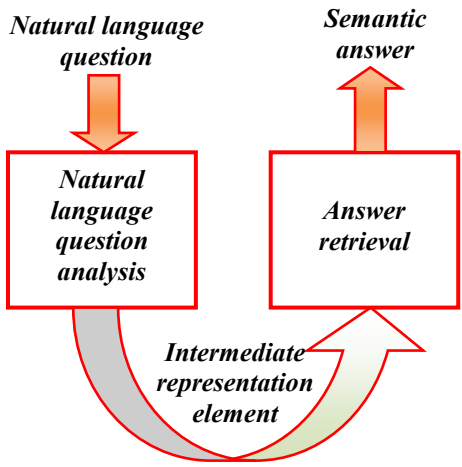

Figure 1. General system architecture.

#### A. Natural language question analysis component

The question analysis component consists of three modules: preprocessing, syntactic analysis and semantic analysis, see Figure 2. It takes the user question as an input and returns a query-tuple representing the question in a compact form. The role of this intermediate representation is to provide an easy way to process the input.

The intermediate representation consists of a "*question-structure*" and one or more query-tuple in the following format:

(*question-structure, question-class, Term1, Relation, Term2, Term3*)

where *"Term1"* represents a concept (object class), *"Term2"* and *"Term3"*, if exist, represent entities (objects), *"Relation"* (property) is a semantic constraint between terms in the question. This representation is meant to capture the semantic of the question.

Simple questions only have one query-tuple and its *question-structure* is the query-tuple's *question-structure*. More complex questions such as composite questions have several sub-questions and each sub-question is represented by a separate query-tuple and the *question-structure* captures this composition attribute.

Our system makes use of GATE [7] infrastructure to analyze questions in natural language. Communication between our system and GATE takes place via the standard GATE API. Existing linguistic processing modules for Vietnamese such as Word Segmentation, Part-of-speech tagger are wrapped as GATE plug-ins. Results of the modules are annotations capturing information such as sentences, words, nouns and verbs. Each annotation has a set of feature-value pairs. For example, a word has a feature *category* storing its part-of-speech tag. This information can then be reused for further processing in subsequent modules. New modules are specifically designed to handle Vietnamese questions using patterns over existing linguistic annotations. This is achieved using GATE JAPE (Java Annotation Pattern Engine) transducers, a set of Jape grammar. A Jape grammar allows one to specify regular expression pattern based on semantic annotations.

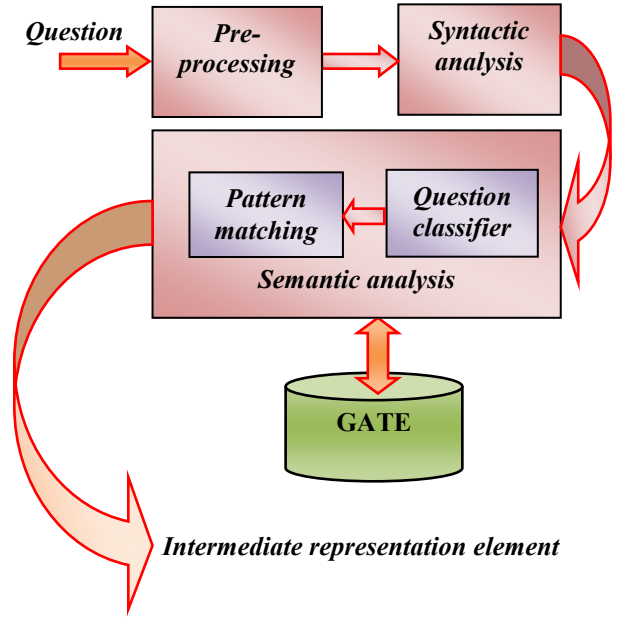

Figure 2. Modules in natural language question analysis component.

*1) Preprocessing Module.* The preprocessing module generates *TokenVn* annotations representing a Vietnamese word with features such as part-of-speech. Vietnamese is a monosyllabic language; hence a word may contain more than one token.

However the Vietnamese word segmentation module is not trained for question domain. There are question phrases, which are indicative of the question categories such as *"phải không"*, tagged as multiple *TokenVn* annotations. In this module we identify those phrases and mark them as single annotations with corresponding feature "*question-word*" and its semantic categories such as: *HowWhy, YesNo$_{trueorflase}$, What, When$_{time}$, Where$_{location}$, Many$_{number}$, Who$_{person}$*. In fact this information will be used to identify the category of a question at a later stage.

*2) Syntactic analysis module.* This module is responsible for identifying noun phrases and the relations between noun phrases. The different modules communicate through the annotations, for example this module uses the *TokenVn* annotations, which is the result of the preprocessing module.

Concepts and entities are normally expressed as noun phrases. Therefore, it is important that we can reliably detect noun phrases in order to generate the query-tuple. We use Jape grammar to specify a pattern over annotations as shown in Table I.

TABLE I. NOUN PHRASE JAPE PATTERN

| | |
|---|---|
| *({TokenVn.category == "Pn"})?* | Quantity pronoun |
| *({TokenVn.category == "Nu"} \| {TokenVn.category == "Nn"})?* | Concrete noun<br>Numeral noun |
| *( {TokenVn.string == "cái"} \| {TokenVn.string == "chiếc"} )?* | "cái$_{the}$"<br>"chiếc$_{the}$" |
| *({TokenVn.category == "Nt"})?*<br>*(*<br>*{TokenVn.category == "Nc" } \|*<br>*{TokenVn.category == "Ng" } \|*<br>*{TokenVn.category == "Na" } \|*<br>*{TokenVn.category == "Np"}*<br>*) +* | Classifier noun<br><br>Countable noun<br>Collective noun<br>Abstract noun<br>Proper noun |
| *({TokenVn.category == "Aa"} \| {TokenVn.category == "An"})?* | Quality adjective<br>Quantity adjective |
| *({TokenVn.string == "này"} \|*<br>*{TokenVn.string == "kia"} \|*<br>*{TokenVn.string == "ấy"} \|*<br>*{TokenVn.string == "đó"} )?* | "này$_{this;\ these}$"<br>"kia$_{that;\ those}$"<br>"ấy$_{that;\ those}$"<br>"đó$_{that;\ those}$" |

When a noun phrase is matched, an annotation *Cumdanhtu*$_{NounPhrase}$ is created to mark up the noun phrase.

The next step is to identify relations between noun phrases or noun phrases and question-words identified by the preprocessing module. After analyzing a number of questions, we use the following four patterns to identify relation phrases:

{Verb}+{Noun Phrase}{Preposition}{Verb}?
{Verb}+{Preposition}?{Verb}?
({"có$_{have|has}$"} | {Verb}){Adjective}{Preposition}{Verb}?
{"có$_{have|has}$"}({Noun Phrase } | {Adjective}){"là$_{is}$"}

When a phrase is matched by one of the relation patterns, an annotation *Moiquanhe*$_{Relation}$ is created to markup the relation. For example, with the following question:

*"ai là sinh viên của lớp khoa học máy tính?"*
*"which student is in the class computer science?"*

The phrase *"là sinh viên của$_{is\ student\ of}$"* is the relation phrase linking the noun phrase *"lớp khoa học máy tính$_{computer\ science\ class}$"* and the *question-word "ai$_{who}$"*.

*3) Semantic analysis module.* The semantic analysis module identifies the question category and the query-tuples that are the result of two sub-modules respectively: the question classifier and the pattern matcher. The output of this module is the intermediate representation of the input question.

*a) Question classifier module.* Question category is indicative of the answer type. It also guides the answer retrieval module. In our system, a question is classified into one of the following ten classes: *HowWhy, YesNo, What, When, Where, Who, Many, ManyClass, List and Entity*.

To identify question categories, we specify a number of Jape grammars using noun phrases annotations and the *question-word* information identified by the preprocessing module. Obviously using this method will result in ambiguity when a question belongs to multiple categories. We allow for this and resolve the ambiguity in the pattern matching module.

*b) Pattern matching module.* This module identifies the question structure and produces the query-tuples as the intermediate representation:

(*question-structure, question-class, Term1, Relation, Term2, Term3*)

This representation is chosen so that it can represent various types of question. Therefore, some terms or relation in the tuple can be missing.

Existing noun phrase annotations and relation annotations are potential candidates for terms and relations respectively. We use Jape grammars to detect the question structure and corresponding terms and relations. We define the following question structures: *"Normal", "Unknterm", "Unknrel", "Definition", "Compare", "ThreeTerm", "Clause", "Combine", "And", "Or", "AffirmNeg_3Term", "AffirmNeg_2Triple", "AffirmNeg"*.

For example, a question has *"Normal"* question structure if it has only one query-tuple and *"Term3"* is missing. Composite questions such as:

*"Danh sách tất cả các sinh viên có quê ở Hà Tây mà học lớp khoa học máy tính?"*
*"List of all students who come from Hà Tây and study in computer science class"*

has question structure of type *"And"* with two query-tuples where **?** represents a missing element:

(*Normal, List, sinh viên$_{student}$, có quê ở$_{come\ from}$, HàTây, ?*)
(*Normal, List, sinh viên$_{student}$, học$_{study}$, lớp khoa học máy tính$_{computer\ science\ class}$, ?*)

## B. Answer retrieval component

The answer retrieval component includes two main modules: Ontology Mapping and Answer Extraction as shown in Figure 3. It takes an intermediate representation produced by the question analysis component and an ontology as its input to generate a semantic answer.

The task of the Ontology Mapping module is to map terms and relations in the query-tuple to concepts, instances and relations in the ontology by using string names. If exact match is not possible, we use string distance algorithm [11][14][15] to find near-matched elements in the Ontology with the similarity measure above a certain threshold. In case ambiguity is still present, the system interacts with the users by presenting different options to get the correct ontology element.

For each query-tuple, the result of the Mapping Ontology module is an ontology-tuple where the terms and relations in the query-tuple are now their corresponding elements in the ontology.

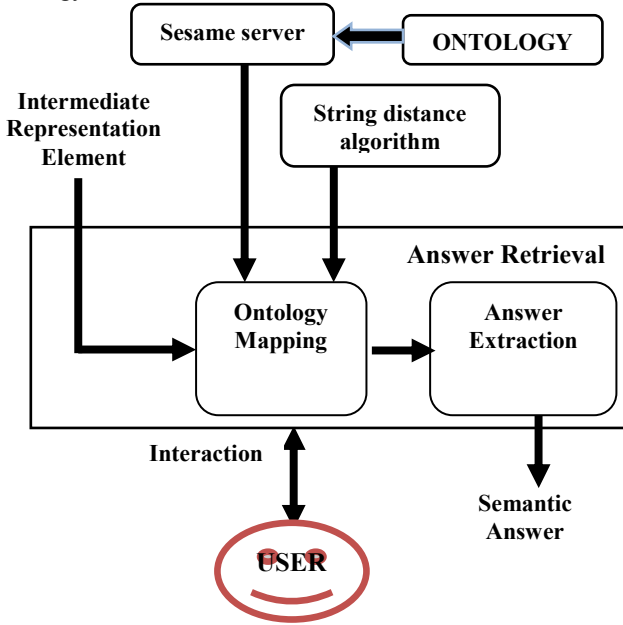

Figure 3. Answer Retrieval Module Architecture.

How the Mapping Ontology module finds corresponding elements in the ontology depends on the question structure. For example, when the query-tuple contains *term1*, *term2* and *relation* with *term3* missing, the mapping process follows the diagram shown in Figure 4. It first tries to match *term1* and *term2* with concepts or instances in the ontology. After that, the set of potential relations in the ontology contains only relations between the two mapped concepts/instances. The ontology relation is then identified in a similar manner as mapping term to a concept or an instance. With the ontology-tuple, the Answer Extraction module find all individuals of the corresponding ontology concept of *term1*, having the ontology relation with the individual corresponding to *term2*. Depending on the question structure and question category, the best semantic answer will be returned.

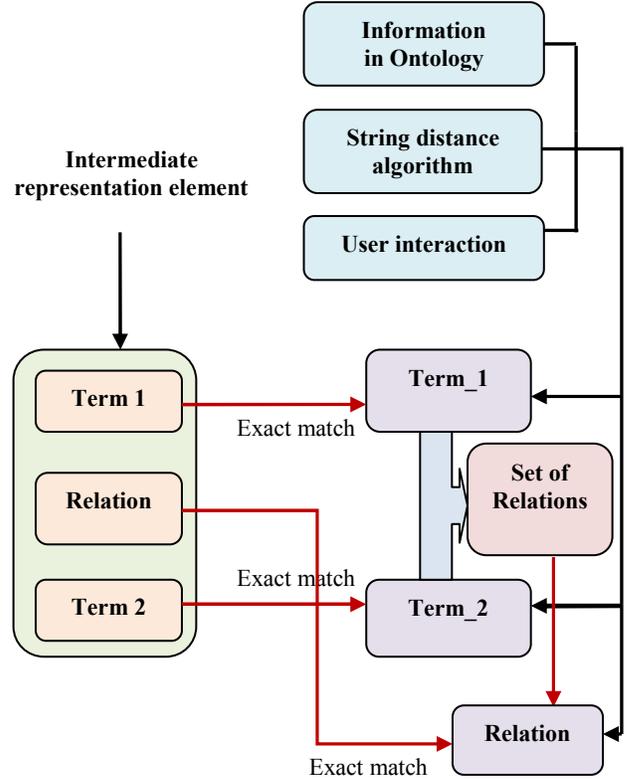

Figure 4. Mapping Ontology module for query-tuple with two terms and a relation.

## IV. ILLUSTRATIVE EXAMPLES

For demonstration and evaluation purposes, we built an ontology containing 15 concepts, such as: *"trường$_{school}$"*, *"giảng viên$_{lecturer}$"*, *"sinh viên$_{student}$"*, 17 attributes or relations, such as: *"học$_{study}$"*, *"giảng dạy$_{teach}$"*, *"là sinh viên của$_{is\ student\ of}$"*, and 78 instances on organizational structure at College of Technology, Vietnam National University Hanoi (see *Appendix A*).

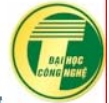

Figure 5. An illustrative example of the system in action.

Consider the question:

*"có bao nhiêu sinh viên học lớp k50 khoa học máy tính?"*

*"how many students studying computer science class k50?"*

The Question analysis component will translate the question into the following query-tuple:

(Normal, ManyClass, sinh viên$_{student}$, học$_{study}$, lớp k50 khoa học máy tính$_{computer\ science\ class\ k50}$, ?)

which is mapped to the following ontology-tuple by the Ontology Mapping module:

(sinh_viên, học, k50_khoa_học_máy_tính)

The Answer Extraction module will in turn generate an answer as shown in Figure 5.

Consider another question, which has a more complex structure:

*"sinh viên nào học lớp k50 khoa học máy tính và có quê ở Hà Nội?"*

*"Which students study at computer science class k50 and come from Hanoi?"*

The Question analysis module determines that this question has a query structure of type "And" with two query-tuples:

(Normal, Entity, sinh viên$_{student}$, học$_{study}$, lớp k50 khoa học máy tính$_{computer\ science\ class\ k50}$, ?)

(Normal, Entity, sinh viên$_{student}$, có quê ở$_{come\ from}$, Hà Nội, ?)

Each query-tuple is then mapped to an ontology tuple by the Ontology Mapping module:

(sinh_viên, học, k50_khoa_học_máy_tính)

(sinh_viên, có_quê_ở, hà_nội)

With each ontology-tuple, the Answer Extraction module find all satisfied individuals in the ontology and a semantic answer is generated based on the question structure *"And"* and the question category *"Entity"*. The returned result is shown in Figure 6.

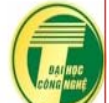

Figure 6. An illustrative example of the system in action.

## V. EXPERIMENT

To evaluate the Question Analysis component, we create a list of 60 questions with varying structures on the domain of College of Technology, Vietnam National University Hanoi. Out of these questions, 57 questions were correctly processed by the Question Analysis component resulting in 95% accuracy. The 5% errors are due to the lack of coverage of our Jape grammars in the pattern-matching module.

Among the 57 questions correctly analyzed, we choose 50 questions with different query-tuples and take their corresponding query-tuples as the set of input to evaluate the Answer Retrieval component. The result is shown in Table II.

TABLE II. QUESTIONS SUCCESSFULLY ANSWERED

| Type | Number question | Percent |
|---|---|---|
| No interaction with users | 25 | 50% |
| With interactions with users | 10 | 20% |
| Number questions successfully answered | 35 | 70% |

Our system gives correct answers to 35 questions. Out of those, 25 questions can be answered automatically without interaction with the user.

Most of the errors in the Answer Retrieval component are due to the Ontology Mapping module as can be seen from Table III. It is mainly because specific terms or relations in the intermediate representation cannot be mapped or incorrectly mapped to corresponding elements in the target ontology.

TABLE III. QUESTIONS WITH UNSUCCESSFUL ANSWERS

| Type | Number question | Percent |
|---|---|---|
| Ontology Mapping errors | 10 | 20% |
| Answer Extraction errors | 5 | 10% |
| Number unsuccessfully answered questions | 15 | 30% |

## VI. CONCLUSION AND FUTURE WORKS

In this paper, we introduce an ontology-based question answering system for Vietnamese. Our system consists of two components: the natural language question analysis and the answer retrieval. We propose an intermediate representation that can capture the semantic structure of the input question, facilitating the processing of matching with the target ontology to find the answer. To the best of our knowledge, this is the first ontology-based question answering system for Vietnamese.

Experimental results of the system on a wide range of questions are promising. Specifically, the Question Analysis module and the Answer Retrieval module achieve an accuracy of 95% and 70% respectively. In the future, we will extend our grammar to provide better coverage for the Question Analysis component and improve the Ontology Mapping module.

ACKNOWLEDGEMENT

This work is partly supported by the research project No. QC.09.08 granted by Vietnam National University, Hanoi.

REFERENCES

[1] A. Galea, "Open-domain Surface-Based Question Answering System", In Proceedings of CSAW, 2003.
[2] A. Popescu, O. Etzioni and H. Kautz, "Towards a theory of natural language interfaces to databases", In Proceedings of IUI, 2003.
[3] A. Saxena, G. Sambhu, S. Kaushik and L. Subramaniam, "IITD-IBMIRL System for Question Answering Using Pattern Matching, Semantic Type and Semantic Category Recognition", In Proceedings of TREC 2007.
[4] B. Katz, G. Borchardt and S. Felshin, "Natural Language Annotations for Question Answering", In Proceedings of FLAIRS 2006.
[5] C. Antonio, F. Francesco, S. Maria and T. Ro, "PIQASso: Pisa question answering system", In Proceedings of TREC 2001.
[6] D. Damljanovic, V. Tablan and K. Bontcheva, "A text-based query interface to owl ontologies", In Proceedings of LREC 2008.
[7] H. Cunningham, D. Maynard, K. Bontcheva, V. Tablan, C. Ursu, M. Dimitrov, M. Dowman, N. Aswani, I. Roberts, Y. Li, A. Sharin and A. Funk, "Developing Language Processing Components with GATE", University of Sheffield, UK. http://gate.ac.uk/.
[8] I.Indroutsopoulos, G. D. Ritchieand P. Thanisch, "Natural Language Interfaces to Databases - An Introduction", In: Natural Language Engineering, 1(1): 29-81, 1995.
[9] K. Nguyen and H. Le, "Natural Language Interface Construction Using Semantic Grammars", In Proceedings of PRICAI 2008.
[10] L. Hirschman and R. Gaizauskas, "Natural Language Question Answering: The View from Here", In: Natural Language Engineering, 7(4):275-300, 2001.
[11] M.Vargas-Vera, E. Motta, "An Ontology-Driven Similarity Algorithm", Technical Report, kmi-04-16, Knowledge Media Institute, The Open University, 2004.
[12] R. Basili, D.H. Hansen, P. Paggio, M.T. Pazienza and F.M. Zanzotto, "Ontological resources and question data in answering", In Proceedings of the Workshop on Pragmatics of Question Answering, USA, 2004.
[13] P. Paggio, D.H. Hansen, R. Basili, M.T. Pazienza and F.M. Zanzotto, "Ontology-based question analysis in a multilingual environment: the MOSES case study", In Proceedings of OntoLex 2004.
[14] V. Lopez, V. Uren, E. Motta and M. Pasin, "AquaLog: An ontology-driven question answering system for organizational semantic intranets", In: Journal of Web Semantics, 5(2):72-105, Elsevier, 2007.
[15] W.W. Cohen, P. Ravikumar, S.E. Fienberg, "A Comparison of String Distance Metrics for Name-Matching Tasks", In Proceedings of the IJCAI-2003 Workshop on Information Integration on the Web, 2003.

APPENDIX A. COLTECH – VNU ONTOLOGY

**CLASS BROWSER**

For Project: school

Class Hierarchy

- owl:Thing
  - trường (2)
    - bộ_môn (5)
    - khoa (3)
    - lớp (5)
    - môn (14)
    - person
      - giảng_viên (4)
      - sinh_viên (15)
    - phòng_thí_nghiệm (1)
  - which
    - chức_vụ (11)
    - học_hàm (2)
    - học_vị (2)
    - mã (3)
    - quê (8)
    - địa_chỉ (3)

**INSTANCE BROWSER**

For Class: sinh_viên

Asserted | Inferred

Asserted Instances
- bùi_tuấn_anh
- nguyễn_bá_đạt
- nguyễn_quốc_đại
- nguyễn_quốc_đạt
- nguyễn_trần_ngọc_linh
- nguyễn_văn_c
- nguyễn_văn_d
- nguyễn_văn_e
- nguyễn_văn_huy
- nguyễn_văn_thanh
- nguyễn_đức_quang
- nguyễn_đức_trung
- phạm_đức_đăng
- trần_bình_giang
- đặng_ngọc_tuyền

Asserted Types
- sinh_viên

**PROPERTY BROWSER**

For Project: school

Object | Datatype | Annotation | All

Properties
- có_bộ_môn_là
- có_chức_vụ_là
- có_hiệu_trưởng_là
- có_học_vị_là
- có_khoa_là
- có_lớp_là
- có_lớp_trưởng_là
- có_mã_là
- có_quê_ở
- có_sinh_viên_là
- có_trưởng_phòng_là
- có_tên_là
- có_địa_chỉ_là
- giảng_dạy
- học
- là_quê_của
- thuộc

Super Properties